\documentclass[sigconf,natbib=false]{acmart}

\AtBeginDocument{%
  }

\copyrightyear{2025}
\acmYear{2025}
\setcopyright{acmlicensed}\acmConference[MM '25]{Proceedings of the 33rd ACM International Conference on Multimedia}{October 27--31, 2025}{Dublin, Ireland}
\acmBooktitle{Proceedings of the 33rd ACM International Conference on Multimedia (MM '25), October 27--31, 2025, Dublin, Ireland}
\acmDOI{10.1145/3746027.3755866}
\acmISBN{979-8-4007-2035-2/2025/10}

\RequirePackage[
  datamodel=acmdatamodel,
  style=acmnumeric,
  ]{biblatex}

\usepackage{xcolor}
\usepackage{multirow}
\usepackage{diagbox}
\usepackage{colortbl}
\usepackage{subcaption}

\begin{document}

\title{TAP: Parameter-efficient Task-Aware Prompting for Adverse Weather Removal}

\author{Hanting Wang}
\orcid{0009-0003-4032-9257}
\affiliation{
  \institution{Zhejiang University}
  \city{Hangzhou}
  \country{China}
}
\email{hantingwang@zju.edu.cn}

\author{Shengpeng Ji}
\orcid{0000-0002-0129-4843}
\affiliation{
    \institution{Zhejiang University}
    \city{Hangzhou}
    \country{China}
}
\email{shengpengji@zju.edu.cn}

\author{Shulei Wang}
\orcid{0009-0009-2285-3606}
\affiliation{
    \institution{Zhejiang University}
    \city{Hangzhou}
    \country{China}
}
\email{shuleiwang@zju.edu.cn}

\author{Hai Huang}
\orcid{0009-0003-8813-2306}
\affiliation{
    \institution{Zhejiang University}
    \city{Hangzhou}
    \country{China}
}
\email{haihuangcode@outlook.com}

\author{Xiao Jin}
\orcid{0000-0002-9227-4011}
\affiliation{%
    \institution{Zhejiang University}
    \city{Hangzhou}
    \country{China}
}
\email{jin.xiao@zju.edu.cn}

\author{Qifei Zhang}
\orcid{0009-0001-8247-4562}
\authornotemark[1]
\affiliation{
    \institution{Zhejiang University}
    \city{Hangzhou}
    \country{China}
}
\email{cstzhangqf@zju.edu.cn}

\author{Tao Jin}
\orcid{0000-0003-3564-1628}
\authornote{Corresponding authors}
\affiliation{
    \institution{Zhejiang University}
    \city{Hangzhou}
    \country{China}
}
\email{jint_zju@zju.edu.cn}

\renewcommand{\shortauthors}{Hanting Wang et al.}

\begin{abstract}
    Image restoration under adverse weather conditions has been extensively explored, leading to numerous high-performance methods. In particular, recent advances in All-in-One approaches have shown impressive results by training on multi-task image restoration datasets. However, most of these methods rely on dedicated network modules or parameters for each specific degradation type, resulting in a significant parameter overhead. Moreover, the relatedness across different restoration tasks is often overlooked. In light of these issues, we propose a parameter-efficient All-in-One image restoration framework that leverages task-aware enhanced prompts to tackle various adverse weather degradations.Specifically, we adopt a two-stage training paradigm consisting of a pretraining phase and a prompt-tuning phase to mitigate parameter conflicts across tasks. We first employ supervised learning to acquire general restoration knowledge, and then adapt the model to handle specific degradation via trainable soft prompts. Crucially, we enhance these task-specific prompts in a task-aware manner. We apply low-rank decomposition to these prompts to capture both task-general and task-specific characteristics, and impose contrastive constraints to better align them with the actual inter-task relatedness. These enhanced prompts not only improve the parameter efficiency of the restoration model but also enable more accurate task modeling, as evidenced by t-SNE analysis. Experimental results on different restoration tasks demonstrate that the proposed method achieves superior performance with only $2.75M$ parameters.
\end{abstract}




\maketitle

\begin{figure}[ht]
\centering
\includegraphics[width=1.0\linewidth]{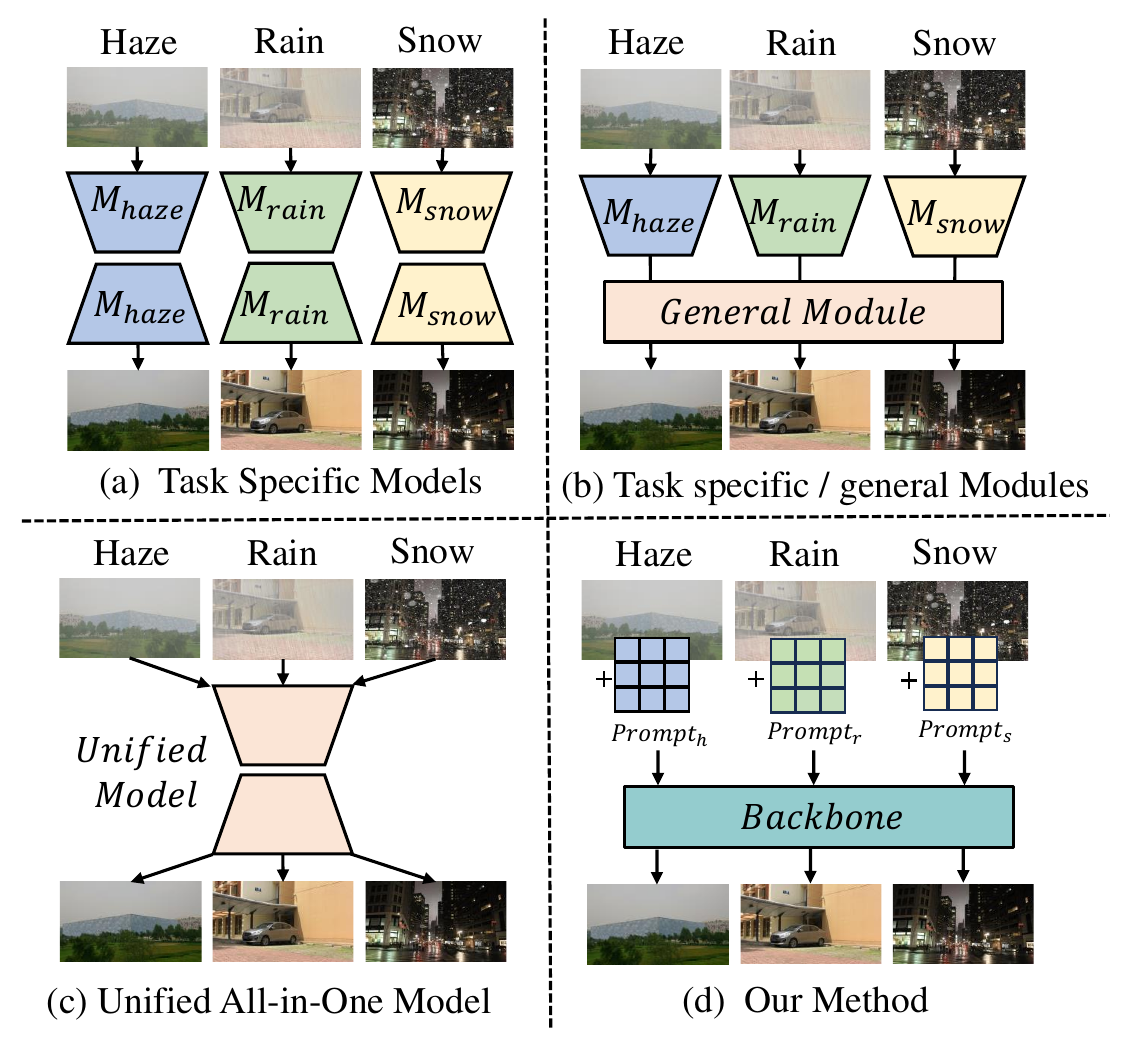}
\caption{High-level view of different image restoration methods.
(a) Training separate models for each task.
(b) Designing task-specific modules or parameters.
(c) Using a unified model to handle all tasks.
(d) Our proposed method. 
Compared to designing dedicated modules for each task or adopting a unified model, our approach achieves a balance between task modeling and parameter efficiency.}
\label{Fig:highview}
\end{figure}

\section{Introduction}
\label{sec:intro}
Adverse weather conditions, such as rain, snow and haze, often reduce the quality of the images during acquisition, which primarily deteriorates the performance of many vision applications. Removing adverse weather conditions from degraded images has been studied intensively, including deraining ~\cite{heavyrain, 2017CNNderain, deraintransformer}, dehazing \cite{2020CNNdehaze, densitydehaze, contrastiveDehaze, dehazeformer}, desnowing \cite{desnownet, snowformer, 2021desnow} and adherent raindrops removal \cite{raindrop2018,adherentraindrop}. Although these approaches have achieved promising results, they are not generic solutions for all adverse weather removal problems, as they are only designed to remove certain typical weathers. However, handling various kinds of weather is inevitable in real-world applications. Such approaches are therefore not capable of satisfying practical needs.

To this end, several methods \cite{transweather,wgwsnet,chen2022learning,promptIR,adair,Histoformer,loraIR,MPerceiver} propose using a single set of network parameters to tackle different types and levels of degradation, commonly referred to as \textit{All-in-One Image Restoration} (AiOIR). As a typical multi-task problem, AiOIR encounters several critical challenges:
\textbf{(I) Task Relatedness.} 
Different restoration tasks may share similar feature patterns while also exhibiting task-specific ones, underscoring the need to model both task-general and task-specific characteristics to achieve better performance.
\textbf{(II) Task Conflict.} During multi-task joint training, the learning of features beneficial for one task might negatively impact another, causing optimization difficulties and performance degradation.
\textbf{(III) Parameter Efficiency.} Achieving optimal performance typically involves introducing task-specific parameters, inevitably leading to increased model complexity and parameter overhead.
These inherent challenges underscore the difficulty in designing a unified model that performs robustly across diverse restoration scenarios.

Existing AiOIR approaches typically balance these challenges through strategic trade-offs in model architectures or training schemes, as shown in Figure \ref{Fig:highview}. One category of methods \cite{allinone,wgwsnet} explicitly incorporates task-specific modules or parameters, thus effectively capturing task-specific features. However, this strategy significantly increases the parameter burden, especially when a larger number of tasks are involved. In contrast, another direction \cite{transweather,Histoformer} seeks to drastically reduce the number of parameters by directly adopting a unified end-to-end architecture. However, due to the lack of explicit modeling of task relatedness, the model often fails to identify the task-shared and task-specific features across tasks, leading to suboptimal performance. To mitigate the drawbacks of both approaches, some efforts \cite{promptIR,MPerceiver,multiExpert} attempt to jointly optimize a shared backbone along with task-related auxiliary modules, aiming to balance task relevance modeling and parameter efficiency. However, the strategy of jointly optimizing the backbone network and task-related modules introduces task conflicts during training, which in turn leads to suboptimal performance. In summary, although existing AiOIR methods have explored various trade-offs between parameter efficiency and task adaptability, none has yet achieved a well-balanced solution to all three challenges. These observations motivate us to propose a novel AiOIR framework that achieves efficient modeling of task relatedness while maintaining parameter efficiency and mitigating task conflicts.

To this end, we introduce a parameter-efficient AiOIR framework, dubbed \textbf{\textit{TAP}}, designed to tackle All-in-One adverse weather degradation removal. As shown in Figure \ref{Fig:highview}, we employ task-specific prompts to strike a balance between task relatedness modeling and parameter efficiency. Compared with integrating task-specific modules (such as encoders \cite{allinone} or expert networks \cite{multiExpert}) into the restoration model, our approach offers greater advantages in terms of parameter efficiency and flexibility. Moreover, in contrast to existing prompt-based image restoration solutions, we introduce task-level trainable soft prompts instead of relying on prompt generators \cite{promptIR} or descriptive texts \cite{instructir} to construct image-specific prompts, enabling more accurate task-level modeling.

To avoid task conflicts during the optimization phase, we adopt a two-stage training strategy consisting of pretraining and prompt tuning. In the pretraining stage, we train the restoration backbone in a supervised manner to learn general restoration knowledge. After this stage, we obtain a network capable of performing coarse image restoration. In the subsequent prompt-tuning stage, we freeze the backbone and optimize only the learnable task-specific prompts to adapt the model to specific tasks. Such a decoupled training strategy effectively mitigates inter-task conflicts that typically arise during joint training (shown in Section \ref{sec:Ablation-train}).

To further facilitate task relatedness modeling, we propose a novel task-aware interactive enhancement strategy for the prompts. 
Considering that tasks exhibit both task-general and task-specific characteristics, we perform a low-rank decomposition of the prompt vectors, using a shared low-rank matrix and individual matrices to model them separately. We refer to this as \textit{implicit interaction enhancement}.
Moreover, we observe that different tasks may exhibit explicit inter-task relationships. For example, certain tasks may share similar feature distributions or restoration patterns, indicating high correlation, while others may involve conflicting objectives or dissimilar degradation distributions, suggesting low relatedness. Previous works \cite{promptIR,multiExpert,wgwsnet} have largely overlooked the modeling of such inter-task relationships. To address this issue, we apply contrastive learning to the decomposed prompts, constraining the task-specific prompt matrices based on the explicit relationships exhibited by different tasks, thereby achieving effective modeling of such inter-task correlations. We refer to this as \textit{explicit interaction enhancement}.
Through the aforementioned task-aware enhancements, we significantly improve the task modeling capability of the prompts, thereby effectively boosting the overall performance of the model.

The key contributions of this paper are as follows:
\begin{enumerate}
    \item We propose a novel task-aware prompting framework for All-in-One Image Restoration in adverse weather removal, achieving a balance between task relatedness modeling and parameter efficiency.
    \item To mitigate task conflicts during the training phase, we propose a two-stage training strategy. We first jointly optimize a general restoration model, and then adapt it to specific tasks using task-bespoke prompts.
    \item We recognize the inter-task relatedness and enhance the prompts through both implicit and explicit interaction mechanisms. The implicit enhancement models the general-specific characteristics across tasks, while the explicit enhancement captures the observable relatedness patterns. These enhanced prompts enable more precise task modeling and improve the model’s ability to adapt to specific tasks.
    \item We validate the effectiveness of the proposed method across four tasks: image deraining, desnowing, dehazing, and raindrop removal. With only around 2.75M parameters, our method achieves state-of-the-art performance.
\end{enumerate}

\begin{figure*}[t]
\centering
\includegraphics[width=1.0\textwidth]{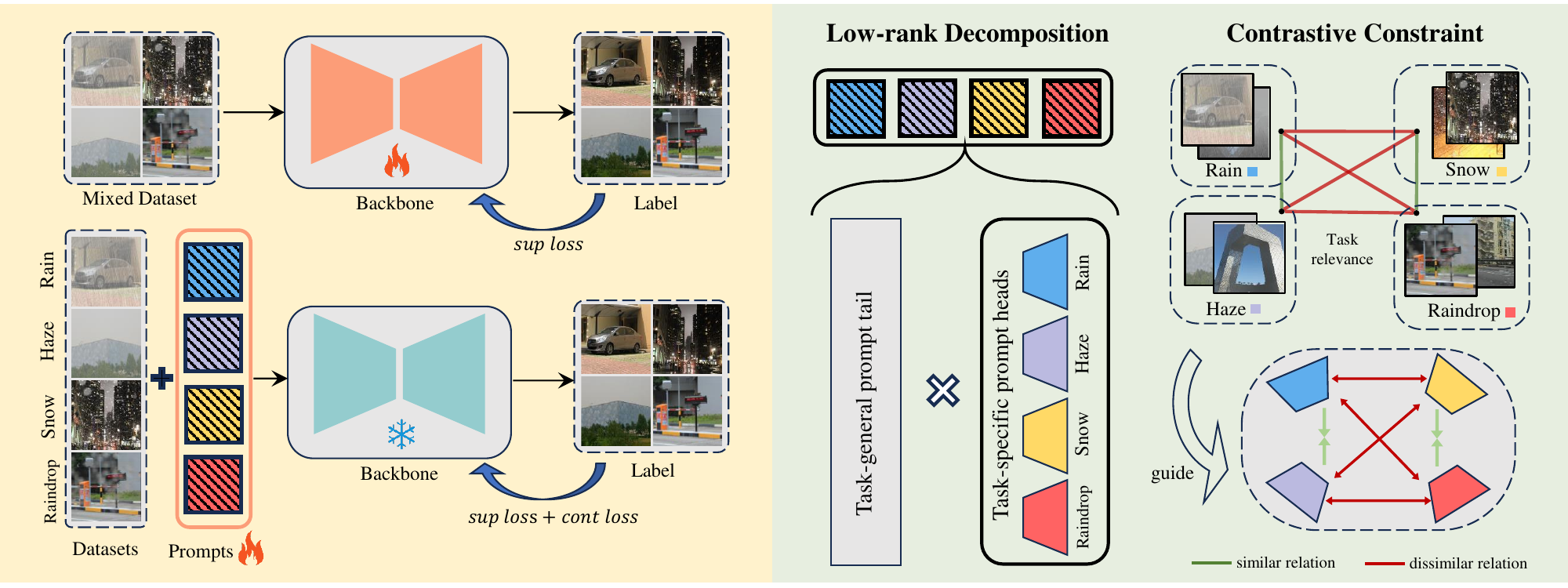} 
\caption{
Overview of the proposed \textit{TAP} framework. The model is first pretrained in a supervised manner and then finetuned with soft prompts. To improve task adaptability, the prompts are enhanced based on inter-task relatedness. Specifically, we apply a low-rank decomposition to factorize the prompt embeddings into a task-shared component and task-specific components (Implicit Interaction Enhancement). Additionally, we impose contrastive constraints on the task-specific heads with reference to inter-task relevance (Explicit Interaction Enhancement).
}
\label{Fig:overall}
\end{figure*}

\section{Related Works}
\label{sec:formatting}

\subsection{All-in-One Image Restoration}
AiOIR methods aim to recover clean images from various types of degradation using a single set of network parameters.
One category of approaches \cite{transweather,Histoformer,dcpt} proposes directly using a unified model to handle all types and levels of degradation simultaneously, typically by training a general image restoration model on mixed-source data.
More mainstream AiOIR methods adopt task-specific modules \cite{allinone,multiExpert} or parameters \cite{wgwsnet} to handle different tasks, achieving effective modeling of specific degradations through explicitly discriminative designs.
Although demonstrating remarkable performance, they also introduce increased computational and memory overhead, limiting their practicality in real-world applications.
In addition, some methods explore multimodal perspectives \cite{daclip,diffweather,instructir,MPerceiver}, leveraging auxiliary modalities such as text to guide the model in handling specific degradations. However, the introduction of extra modalities inevitably leads to cross-modal alignment errors \cite{huang01,huang02,huang03}.
Moreover, generative modeling \cite{ddpm,sde,recitedflow}\nocite{shulei01,linGen01,linGen02,linGen03,ji001,fu01} has attracted growing interest in image restoration \cite{irsde,rddm,ddrm,irbridge}, but its commonly used iterative generation strategies still suffer from higher computational costs compared to supervised methods.

Mainstream approaches reveal a trade-off between task modeling granularity and parameter efficiency.
Inspired by advances in prompt learning \cite{softprompt,lowrankprompt,dept,mobilespeech}, we propose using task-aware prompts for more parameter-efficient and flexible task modeling.

\subsection{Prompt-based Image Restoration}
In recent years, prompt-based image restoration methods have made remarkable progress, including approaches based on visual prompts, textual prompts, and multimodal prompts \cite{allinonesurvey}. 
Visual prompting methods \cite{promptIR,ProRes} are typically designed at the instance level, where customized prompting modules are used to represent degraded images as visual embeddings to guide the restoration model.
However, these prompting modules are usually optimized jointly with the backbone network, making it difficult to avoid parameter conflicts across different tasks. Moreover, the effectiveness of visual prompts is often limited by the semantic gap \cite{segmentgap}.
\nocite{fu01,linUn01,linUn02,linUn03,linUn04,linGen04,linGen05,linUn06,linUn07,linUn08,linUn09}

Textual prompting methods \cite{ldr,textualPromptIR,instructir} leverage natural language to provide user-friendly and semantically precise guidance.
Nevertheless, degradations in images are often hard to accurately abstract into natural language, and cross-modal alignment errors hinder the ability of textual embeddings to effectively guide the model in handling degradations.
Multimodal prompting methods \cite{MPerceiver,daclip,diffweather} utilizing vision-language models (VLMs) take advantage of their powerful cross-modal representation capabilities and demonstrate promising performance. However, the introduction of VLMs significantly increases the parameter burden due to their unignorable model size.

In this work, we propose using task-level prompts to guide the model in adapting to specific restoration tasks. Compared to instance-level visual prompting methods, our approach explicitly incorporates task-related information to guide the model to handle specific degradations, thereby mitigating the distribution bias introduced by individual samples. Moreover, we decouple the optimization of prompts from the backbone network to avoid task parameter conflicts. Additionally, the employed optimization-based soft prompts offer a more precise representation of task compared to directly using textual prompts. This design not only reduces cross-modal alignment errors but also avoids the significant overhead introduced by VLMs.

\section{Method}
\label{sec:Method}
In this section, we introduce the proposed TAP framework. The overview of our method is illustrated in Figure \ref{Fig:overall}.

\subsection{Model Architecture}
\label{sec:ModelArchitecture}
Our model adopts an improved architecture built upon SwinIR~\cite{SwinIR}, which has shown impressive performance across various image restoration tasks. A key motivation for selecting SwinIR as our backbone is its favorable trade-off between restoration accuracy and parameter efficiency, owing to its relatively lightweight design. 

We employ a five-stage U-Net structure where the traditional convolutional blocks are replaced with Transformer-based modules. Inspired by \cite{dehazeformer}, the original concatenation fusion and Layer Normalization are replaced with SK fusion and Rescale Layer Normalization. In addition, some attention layers in the decoder are removed to further reduce the number of parameters and computational complexity. Further details are provided in the supplementary material.

To achieve a unified modeling of adverse weather degradations, we reconstruct the physical degradation model proposed in \cite{heavyrain} and employ a soft reconstruction following \cite{dehazeformer}:
\begin{equation}
    X_{hq} = K \odot X_{lq} + R + X_{lq},
    \label{ourformulation}
\end{equation}
where $\odot$ is element-wise multiplication, $X_{hq}$ and $X_{lq}$ represent the high-quality and low-quality image, respectively. The coefficient terms are defined as $K = \frac{1}{T} - 1, R = \left(1 - \frac{1}{T}\right)A + r$, where $T$ denotes the transmission map caused by scattering, $A$ is the atmospheric light, and $r$ is an additional residual term. Given a degraded image $X_{lq} \in \mathbb{R}^{H\times W \times 3}$ as input, we drive the model to predict a feature map $O \in \mathbb{R}^{H\times W \times 4}$ and then split $O$ into $K \in \mathbb{R}^{H\times W \times 1}$ and $R \in \mathbb{R}^{H\times W \times 3}$ to compute the predicted high-quality image $\hat{X}_{hq}$ using Eq (\ref{ourformulation}).

The core component of the adopted network, which follows the typical Transformer architecture \cite{transformer}, is the multi-head self-attention (MHSA) module.
Given an input feature map $X \in \mathbb{R}^{b \times c \times h \times w}$, the Swin Transformer \cite{swintransformer} applies MHSA by first projecting $X$ into query, key, and value using linear layers, and then partitioning the tokens into non-overlapping windows. Within each window, self-attention is computed as:
\begin{equation}
    \text{Attention}(Q, K, V) = \text{Softmax}(\frac{QK^T}{\sqrt{d}}+B)V,
\end{equation}
where $Q,K,V \in \mathbb{R}^{b' \times l \times d}$ denote the query, key and value respectively, $l$ is the number of tokens per window, $d$ is the dimension, $b'=b \times (h \times w) / l$ and $B \in \mathbb{R}^{l \times l}$ is the relative position bias term. In the subsequent prompt tuning phase, we primarily adjust the MHSA module to adapt the model to specific tasks.

\subsection{Attention-level Soft Prompts}
An intuitive approach \cite{attentionprompt} to introducing prompts is to directly concatenate trainable prompt embeddings with the input hidden states of the attention module. Theoretically, this approach is equivalent to concatenating prompts to the query, key, and value simultaneously. However, such a setup introduces dimensional changes, making trimming of the output representation inevitable. We experiment with this approach and do not observe any significant performance improvements (Section \ref{sec:Ablation-prompt}). Our interpretation is that the trimming of the output representation results in information loss. 
Therefore, we introduce attention-level prompts at the inputs of the MHSA layers.
Suppose the length of the prompts is $m$, the prompting process can be formulated as:
\begin{equation}
    \text{Attention}^*(Q, K, V) = \\ 
    \text{softmax}(\frac{Q[\mathcal{P}_k, K]^T}{\sqrt{d}}+B)[\mathcal{P}_v, V],
    \label{promptAttention}
\end{equation}
where $[\cdot]$ is the concatenation operation, $\mathcal{P}_k,\mathcal{P}_v \in \mathbb{R}^{m \times d}$ denote the prompt vectors inserted into the key $K$ and value $V$, respectively. 
We incorporate prompt embeddings into the key and value of self-attention, thereby obtaining adjusted attention representations. 
This setup enables modification of all dimensions in the output representation.
Additionally, as the prompts do not prepend to the query vector, the output sequence length remains the same as the input sequence. The inclusion of task-specific soft prompts, which are not shared between tasks, serves to adapt the model to process specific degradations. 

\subsection{Task-aware Enhancement of Prompts}
As mentioned before, we improve the task modeling capability of the prompts through task-aware enhancement.
Specifically, we model the task-general and task-specific characteristics through \textit{implicit interaction enhancement}, and capture the observable inter-task relatedness through \textit{explicit interaction enhancement}.

\noindent \textbf{Implicit Interaction Enhancement.}
Low-rank decomposition has proven effective for efficient parameter tuning~\cite{lora}. In the context of prompt learning, prior works~\cite{lowrankprompt,dept} show that applying low-rank decomposition to soft prompts not only reduces parameter count but also serves as a regularizer to alleviate overfitting. 
Beyond these benefits, we further perform low-rank decomposition on the soft prompts to obtain a task-shared component and a set of task-specific components, which enable the modeling of both task-general and task-specific characteristics.

As shown in Figure \ref{Fig:overall}, we formally decompose the prompt vectors as follows:
\begin{equation}
    \mathcal{P}^i = \mathcal{P} ^i_s \times \mathcal{P} _g,
\end{equation}
where $\times$ denotes matrix multiplication, $\mathcal{P}^i_s \in \mathbb{R}^{m \times lr}$ is the task-specific prompt head of the $i$-th weather removal task, $\mathcal{P}_g \in \mathbb{R}^{lr \times d}$ is the task-general prompt tail. Here, $lr$ refers to the rank of the parameter matrix. 
The task-specific prompt heads are specific to each task, while the task-general prompt tail is shared among all tasks. 
In this way, task-specific characteristics are incorporated into the task-specific prompt heads, while general knowledge across tasks is captured by the task-general prompt tail.

\noindent \textbf{Explicit Interaction Enhancement.}
We note that there exists observable relatedness among different restoration tasks. 
In Figure \ref{fig:tsne}(a), we visualize the interactions between the residual features of different weather degradation image pairs using t-SNE \cite{tSNE}. These features are extracted by a pretrained VGG16 model \cite{VGG}.
The visualization shows that snow and raindrop degradations are more similar to each other, while rain and haze also exhibit a degree of similarity.
We speculate that this may be due to snow and raindrop degradation both have similar droplet-shaped occlusions, while rain exhibits stripe-like artifacts and transparency changes similar to haze. 
Previous works \cite{promptIR,wgwsnet} often focus solely on explicitly distinguishing between different tasks, while overlooking the relatedness among tasks, which may lead to suboptimal task modeling.

To model these explicit interaction relationships, 
we propose imposing contrastive constraints on the decomposed task-specific prompt heads to better align them with the actual inter-task relationships.
Our objective is to achieve higher similarity between prompts corresponding to the degradations with closer explicit interaction relationships.
Conversely, we expect lower similarity between prompts corresponding to degradations with weaker explicit interaction relationships.
To this end, we employ a contrastive loss with multiple positives, following \cite{supervisedContrastiveLearning}, as a particular degradation may be similar to multiple other types. Assume there are $N$ types of weather tasks. Let $i \in I \equiv \{1\cdots N\}$ be the index of an arbitrary task, the contrastive loss is formulated as:
\begin{equation}
    \begin{aligned}
    &\mathcal{L}_{contrastive} = \sum_{i \in I}\mathcal{L}_{contrastive}^{i} \\
     = & \sum_{i \in I}-\frac{1}{\vert T^+_{(i)} \vert}\sum_{p \in T^+_{(i)}}\log\frac{\exp{\text{sim}(\mathcal{P}_i, \mathcal{P}_{p})}/\tau}{\sum\limits_{k \in I}\mathbf{1}_{i \neq k}\exp{\text{sim}(\mathcal{P}_i, \mathcal{P}_k)}/\tau},
    \end{aligned}
    \label{eq:contrastiveloss}
\end{equation}
where $T^+_{(i)}$ is the set of indices of all tasks that have similar relationships to the $i$-th task distinct from $i$, $\vert T^+_{(i)} \vert$ is its cardinality, $\tau$ is the temperature parameter, $\mathbf{1}_{i \neq j}$ is an indicator function that takes the value 1 when $i \neq j$ and 0 otherwise and $\text{sim}(x, y)$ is the cosine similarity between two vectors $x$ and $y$. This optimization objective essentially serves as a regularization, which we use to constrain the similarity among prompts for better alignment with the explicit relationships across tasks. Rather than simply enforcing the task-specific prompts to distinguish from one another, we expect them to more accurately guide the model in identifying and addressing the associated degradation artifacts. 

\subsection{Two-Stage Training Strategy}
\label{sec:Method-Train}
To avoid task conflicts caused by joint training from scratch, we explicitly divide the training process into two stages. In the pretraining stage, we employ a shared backbone network to learn the weather-general features in a supervised manner using a mixed dataset from various weather conditions. The loss function of the pretraining stage is formulated as follows:
\begin{equation}
    \mathcal{L}_{pretrain}=\sum_{i\in I}(\Vert \hat{X}_{hq}^i - X_{hq}^i \Vert_1 + \lambda_{per}\sum_{j \in S}(\Vert \varphi_j (\hat{X}_{hq}^i) - \varphi_j (X_{hq}^i) \Vert_2))
\end{equation}
where $\hat{X}_{hq}^i, X_{hq}^i$ respectively denote the predicted high-quality images and ground-truth images of the $i$-th weather, $\lambda_{per}$ is the hyperparameter of the regularization of perceptual loss, $\varphi_{j}$ denotes the $j$-th layer of the pretrained VGG16 model. By default, we set $\lambda_{per} = 0.1$ and $S=\{3, 8, 15\}$ following \cite{transweather}. We also experiment with settings using simple L1 loss and MSE loss but observed slight overfitting (Section \ref{sec:Ablation-train}), consistent with the findings of \cite{SemiUIR}.

In the second finetuning stage, we freeze the parameters of the backbone and train the task-level prompts with L1 loss and contrastive loss, which can be formulated as:
\begin{equation}
    \mathcal{L}_{finetune}=\sum_{i\in I}(\Vert \hat{X}_{hq}^i - X_{hq}^i \Vert_1) + \lambda_{cont}\mathcal{L}_{contrastive}
\end{equation}
where $\mathcal{L}_{contrastive}$ is the contrastive loss defined in Eq (\ref{eq:contrastiveloss}), and $\lambda_{cont}$ is the weighting hyperparameter for the contrastive loss term.

\begin{table*}[t]
\centering
\caption{Quantitative results on four challenging image restoration datasets, cover deraining, desnowing, dehazing and raindrop removal. We report the PSNR $\uparrow$ and SSIM $\uparrow$ results for both task-specific and All-in-One methods. 
}
\scalebox{0.88}{\begin{tabular}{c|l|c|cc|cc|cc|cc|>{\columncolor{gray!20}}c>{\columncolor{gray!20}}c}
\toprule
\multirow{2}{*}{\textbf{Type}} & \multirow{2}{*}{\textbf{Method}} & \multirow{2}{*}{\#Params} 
& \multicolumn{2}{c|}{\textbf{OutdoorRain} \cite{heavyrain}} 
& \multicolumn{2}{c|}{\textbf{Snow100K-L} \cite{desnownet}} 
& \multicolumn{2}{c|}{\textbf{RESIDE-SOTS} \cite{reside}} 
& \multicolumn{2}{c|}{\textbf{RainDrop} \cite{raindrop2018}} 
& \multicolumn{2}{c}{\cellcolor{gray!20} \textbf{Average}} \\
& & & PSNR $\uparrow$ & SSIM $\uparrow$ & PSNR $\uparrow$ & SSIM $\uparrow$ & PSNR $\uparrow$ & SSIM $\uparrow$ & PSNR $\uparrow$ & SSIM $\uparrow$ & PSNR $\uparrow$ & SSIM $\uparrow$ \\
\midrule
\multirow{3}{*}{Task-Specifc}
& SwinIR \cite{SwinIR}        & 12M   & 23.23 & 0.8692 & 28.18 & 0.8803 & 21.50 & 0.8912 & 30.82 & 0.9041 & 25.93 & 0.8862 \\
& MPRNet \cite{mprnet}        & 4M    & 30.25 & 0.9143 & 28.66 & 0.8686 & 24.27 & 0.9124 & 30.99 & 0.9163 & 28.54 & 0.9029 \\
& Restormer \cite{restormer}     & 18M   & 29.22 & 0.9059 & 29.37 & 0.8812 & 24.09 & 0.9271 & 31.21 & 0.9188 & 28.47 & 0.9083 \\
\midrule
\multirow{13}{*}{All-in-One}
& All-in-One \cite{allinone}     & 44M   & 24.71 & 0.8980 & 28.33 & 0.8821 & 30.49 & 0.9498 & 31.12 & 0.9321 & 28.66 & 0.9155 \\
& TransWeather \cite{transweather}   & 38M   & 23.18 & 0.8415 & 27.80 & 0.8534 & 27.66 & 0.9517 & 28.98 & 0.9033 & 26.91 & 0.8875 \\
& TKMANet  \cite{chen2022learning}           & 29M   & 23.94 & 0.8523 & 29.27 & 0.8759 & 30.76 & 0.9725 & 31.81 & 0.9107 & 28.95 & 0.9029 \\
& WGWSNet \cite{wgwsnet}       & 6M    & 25.31 & 0.8967 & 29.71 & 0.8892 & 30.85 & 0.9795 & 31.31 & 0.9297 & 29.80 & 0.9238 \\
& $\text{WeatherDiff}_{64}$ \cite{weatherDiffusion} & 87M   & 29.64 & 0.9312 & 30.09 & 0.9041 & 31.15 & 0.9603 & 30.71 & 0.9312 & 30.40 & 0.9317 \\
& $\text{WeatherDiff}_{128}$ \cite{weatherDiffusion}  & 87M   & 29.72 & 0.9216 & 29.58 & 0.8941 & 31.08 & 0.9598 & 29.66 & 0.9225 & 30.01 & 0.9245 \\
& PromptIR\textsuperscript{†} \cite{promptIR}     & 35M   & 29.84 & 0.9141 & 30.88 & 0.8978 & 30.58 & 0.9742 & 31.12 & 0.8974 & 30.11 & 0.9209 \\
& UtilityIR \cite{alwaysclear}     & 26M   & 31.16 & 0.9267 & 29.47 & 0.8794 & 30.76 & 0.9593 & 32.01 & 0.9253 & 31.88 & 0.9405 \\
& AWRCP \cite{codebookprior}          & -     & 31.39 & 0.9329 & 31.92 & \textbf{0.9341} & -     & -      & 31.93 & 0.9314 & -     & -      \\
& MPerceiver \cite{MPerceiver}    & -     & 31.25 & 0.9246 & 31.02 & 0.9164 & -     & -      & 33.21 & 0.9294 & -     & -      \\
& Histoformer \cite{Histoformer}    & 17M   & 32.08 & 0.9389 & 32.16 & 0.9261 & 30.93 & 0.9758 & 33.06 & 0.9441 & 32.56 & 0.9462 \\
& LoRA-IR \cite{loraIR}       & 85M   & 32.62 & 0.9447 & 32.28 & 0.9296 & 30.68 & 0.9610 & \textbf{33.39} & \textbf{0.9489} & 32.74 & 0.9461 \\
\cline{2-13}
& \textbf{TAP}  & \textbf{2.75M} & \textbf{32.71} & \textbf{0.9493} & \textbf{32.88} & 0.9278 & \textbf{32.77} & \textbf{0.9823} & 32.93 & 0.9416 & \textbf{32.82} & \textbf{0.9503} \\
\bottomrule
\end{tabular}}
\label{tab:comparison}
\end{table*}

\begin{figure*}[t!]
    \centering
    \includegraphics[width=1.0\linewidth]{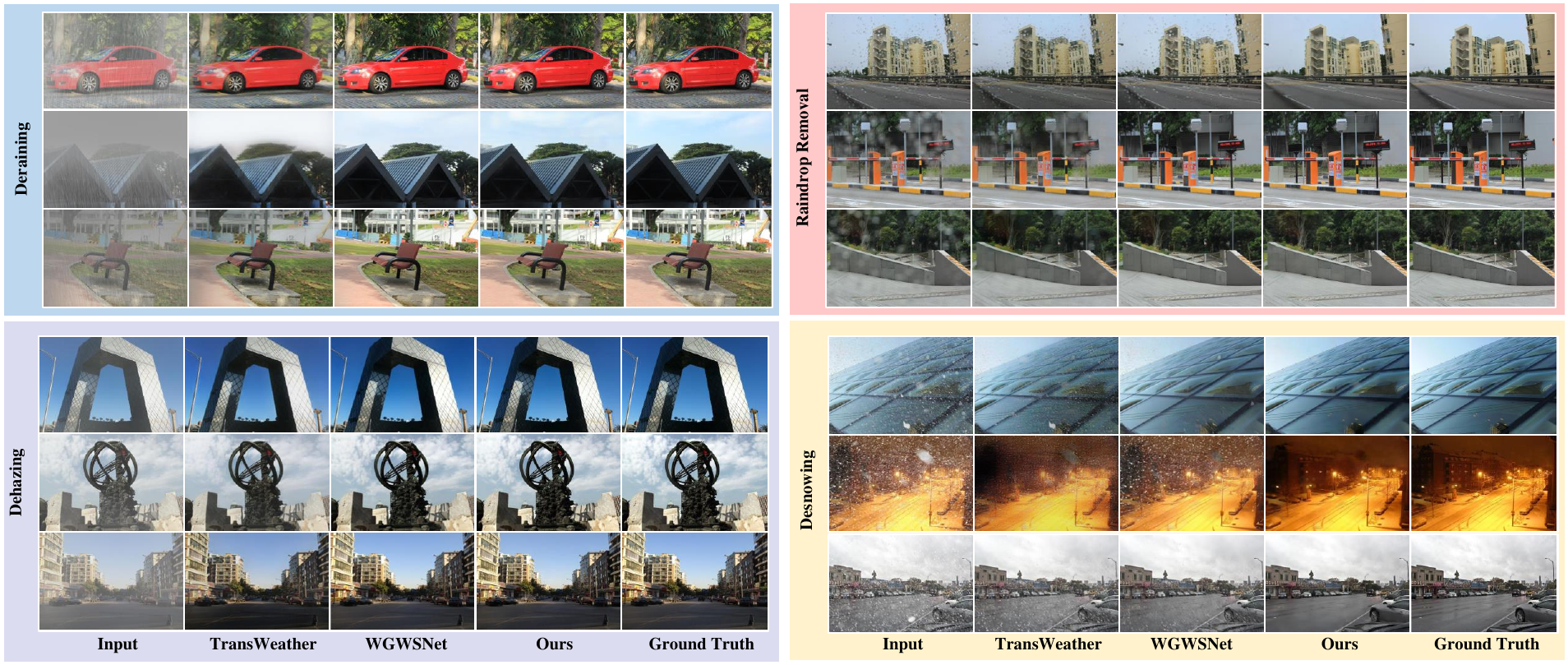}
    \caption{Visual comparison with other all-in-one multi-weather image restoration methods. The proposed TAP achieves better visual quality compared to other AiOIR methods.}
    \label{fig:visualization}
\end{figure*}

\begin{table*}[t]
\centering
\caption{
No-reference IQA results of All-in-One methods on three real-world datasets. We report the results of MUSIQ \cite{musiq}, BRISQUE \cite{brisque}, and NIQE \cite{nique} The evaluation is conducted under a unified setting.
}
\scalebox{0.95}{
\begin{tabular}{c|ccc|ccc|ccc}
\toprule
\textbf{Dataset} & \multicolumn{3}{c|}{\textbf{RealRain-1K} \cite{realrain}} & \multicolumn{3}{c|}{\textbf{RealSnow} \cite{wgwsnet}} & \multicolumn{3}{c}{\textbf{RTTS} \cite{reside}} \\
\cmidrule(r){0-2} \cmidrule(r){2-4} \cmidrule(r){5-7} \cmidrule(r){8-10}
\textbf{Method} & MUSIQ$\uparrow$ & BRISQUE$\downarrow$ & NIQE$\downarrow$ & MUSIQ$\uparrow$ & BRISQUE$\downarrow$ & NIQE$\downarrow$ & MUSIQ$\uparrow$ & BRISQUE$\downarrow$ & NIQE$\downarrow$ \\
\midrule
PromptIR\cite{promptIR} & 33.187 & 46.882 & 12.551 & 39.854 & 40.872 & 10.874 & 45.256 & 52.825 & 9.852 \\
WGWSNet\cite{wgwsnet} & 39.758 & 42.261 & 10.641 & 50.289 & \textbf{31.924} & 4.258  & 47.952 & 53.578 & 11.587 \\
$\text{WeatherDiff}_{64}$\cite{weatherDiffusion} & 41.975 & 41.870 & 9.587  & 47.586 & 33.879 & 6.881  & 49.888 & 51.574 & 10.054 \\
UtilityIR\cite{alwaysclear} & 41.758 & 43.821 & 9.001  & 49.812 & 32.987 & 5.027  & 50.517 & 50.954 & 8.805 \\
LoRA-IR\cite{loraIR} & 43.439 & \textbf{39.843} & 8.393  & 45.812 & 35.871 & 6.871  & 48.109 & 54.641 & 9.847 \\
\midrule
\textbf{TAP} & \textbf{44.915} & 39.915 & \textbf{8.147} & \textbf{51.969} & 32.347 & \textbf{4.002} & \textbf{52.076} & \textbf{41.315} & \textbf{5.003} \\
\bottomrule
\end{tabular}
}
\label{tab:realword}
\end{table*}

\section{Experiments}
\label{sec:experiment}
In this section, we present the results of qualitative and quantitative experiments to validate the effectiveness and superiority of the proposed method.

\subsection{Datasets}
We use synthetic datasets to train our model corresponding to different tasks, including the OutdoorRain dataset \cite{heavyrain} for deraining, the Snow100K dataset \cite{desnownet} for desnowing, the RESIDE dataset \cite{reside} for dehazing, and the RainDrop dataset \cite{raindrop2018} for raindrop removal. During the pretraining stage, to maintain balance across tasks, we ensure that each batch contains an equal number of samples from each task.
At the testing stage, we retain the original dataset settings, including the number of samples and image resolutions, to ensure a fair comparison. More details can be found in the supplementary material. 
In addition, beyond the aforementioned synthetic datasets, we further evaluate the generalization ability of our model on real-world scenarios using RealRain-1K\cite{realrain} for deraining, RealSnow\cite{wgwsnet} for desnowing, and RTTS\cite{reside} for dehazing. The Raindrop dataset \cite{raindrop2018} is collected from real-world scenes, and thus can reflect the model's performance in real-world scenarios.

\begin{figure*}[t]
\centering
\includegraphics[width=1.0\linewidth]{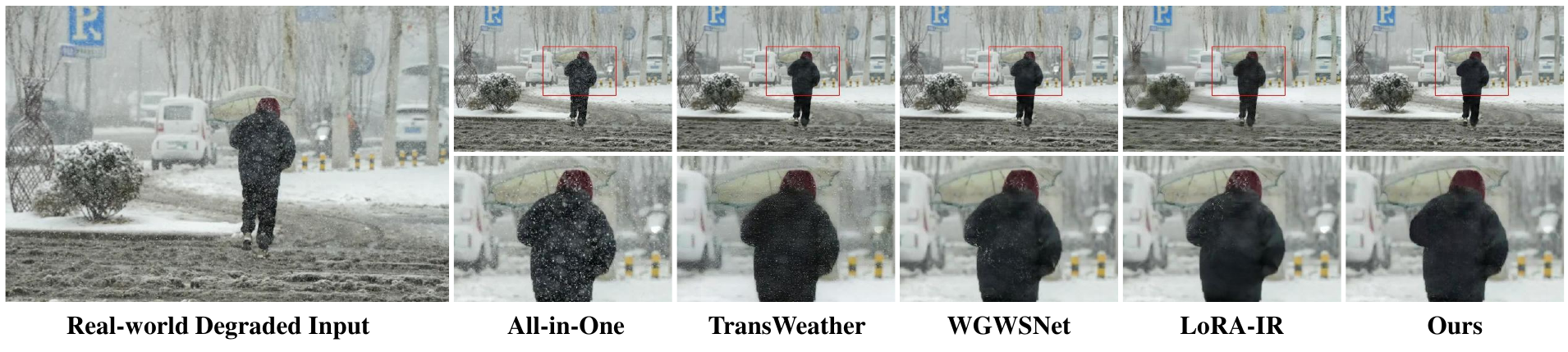} 
\caption{
    Visual results on real-world snow-degraded images. We collected real-world snowy images from the internet to evaluate the generalization ability of different methods in real scenarios. Compared with other approaches, the proposed method removes snow-related artifacts in a more fine-grained manner.
}
\label{Fig:real}
\end{figure*}

\subsection{Implement Details}
We implement our method on the Pytorch platform. We train the model using 4 * NVIDIA RTX 2080Ti GPUs. During the pretraining stage, our model is trained for 200 epochs with a batch size of 32. We use an initial learning rate of $3.0 \times e^{-4}$ which is adjusted with the cosine annealing scheme. During the finetuning stage, we train the soft prompts for 100 epochs with a batch size of 32, while keeping the backbone network parameters frozen. We use an initial learning rate of $5.0 \times e^{-5}$. 
During both training stages, we apply random flipping as a data augmentation strategy and randomly crop the images into patches of size $256 \times 256$.

\subsection{Comparison with State-of-the-Art methods}
We conduct comparative experiments on both synthetic datasets and real-world degraded images to evaluate the proposed method.

\noindent{\bf Quantitative Comparison Results on Synthetic Datasets.}
Table \ref{tab:comparison} presents the quantitative results on four adverse weather removal tasks. 
Compared to the original SwinIR\cite{SwinIR}, our method achieves an average PSNR improvement of 6.89 dB, demonstrating the effectiveness of our architectural improvements. It also outperforms the representative PromptIR\cite{promptIR} by 2.71 dB, validating the advantage of two-stage training in mitigating task conflicts over single-stage joint optimization. Even when compared with the latest LoRA-IR\cite{loraIR}, our approach still achieves a 0.12 dB gain while using only 2.75M parameters, demonstrating its superior performance.

\noindent{\bf Qualitative Comparison Results on Synthetic Datasets.}
Figure \ref{fig:visualization} presents visual comparisons across the four tasks. Compared to other AiOIR methods, the proposed TAP delivers more visually pleasing and satisfactory results.

\noindent{\bf Quantitative Comparison Results on Real-world Datasets.}
Table \ref{tab:realword} presents the quantitative comparison results of the proposed TAP on real-world datasets across three different tasks. Since ground truth is unavailable, we employ no-reference image quality assessment metrics to evaluate performance. Compared to the latest method LoRA-IR, the proposed TAP achieves an average improvement of approximately 16.33\% across all tasks, demonstrating its superior generalization ability in real-world scenarios.

\noindent{\bf Qualitative Comparison Results on Real-world Degraded Images.}
Figure \ref{Fig:real} shows visual comparisons on real-world snow-degraded images. 
The proposed TAP method achieves more effective removal of snow artifacts in human regions than existing approaches, highlighting its enhanced fine-grained restoration capability.

\section{Ablation Study}
\label{sec:Ablation}
We conduct ablation studies to validate the effectiveness of the prompting strategy and length (Section \ref{sec:Ablation-prompt}), task-aware prompt enhancement (Section \ref{sec:Ablation-enhance}), and the training strategy and hyperparameters (Section \ref{sec:Ablation-train}) in this section.

\subsection{Prompting Strategy}
\label{sec:Ablation-prompt}

\noindent\textbf{Attention-level Prompts.} 
Table \ref{tab:ablation} presents the quantitative PSNR results under different prompt strategies, with all prompt lengths set to 12. 
We first evaluate the performance of prompts under different strategy formulations.
Compared to the classical strategy \cite{attentionprompt} that extends learnable vectors to all QKV inputs ('$+ P_{full}$'), our customized attention-level prompts ($+ P_{attn}$) achieves a more substantial improvement (avg. PSNR gain of 0.79 dB). We attribute this to the fact that the former strategy requires discarding the extended output portion, which leads to information loss.

\noindent\textbf{Prompt Length.} 
The diagonal entries of Table \ref{tab:ablationforrank} show the quantitative results with different prompt lengths when the rank is set to 0 (\textit{i.e.}, without low-rank decomposition enhancement). We observe that the performance gain from length 12 to 16 is less significant compared to that from 8 to 12 (0.238 dB vs. 0.013dB). This aligns with the conclusion in \cite{lowrankprompt}, which suggests that the performance of soft prompts initially improves with increasing rank but gradually saturates, indicating a performance ceiling. Therefore, considering the computational cost during training, we adopt a prompt length of 12 as the default setting.

\subsection{Task-Aware Enhancement}
\label{sec:Ablation-enhance}

\begin{figure}[t]
\centering
\includegraphics[width=0.8\linewidth]{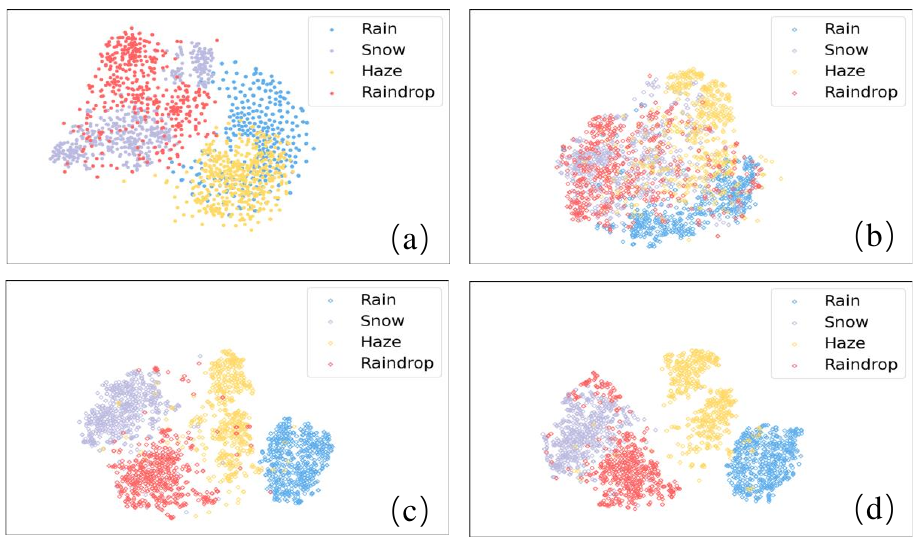} 
\caption{
    Visualization of t-SNE analysis. (a) t-SNE distribution across datasets with different weather types. (b) t-SNE distribution of intermediate features from the pretrained baseline. (c) t-SNE distribution of intermediate features from the model finetuned with task-specific prompts. (d) t-SNE distribution of intermediate features from the model finetuned with task-aware enhanced prompts.
}
\label{fig:tsne}
\end{figure}

\noindent\textbf{Implicit interaction enhanced prompts.} 
As shown in Table \ref{tab:ablationforrank}, when the rank is set to 4, the model consistently outperforms the baseline without low-rank decomposition (diagonal entries). This result suggests the effectiveness of the proposed implicit interaction enhancement. In the prompt tuning phase, the task-general prompt tail models the shared characteristics across tasks, enhancing the ability of the prompts to adapt the model to specific tasks. Notably, we find that the model achieves the best performance when the prompt length is set to 12 and the rank is set to 4. Therefore, we adopt this configuration as the default setting for implicit interaction enhancement.

\begin{figure}[t]
\centering
\includegraphics[width=0.8\linewidth]{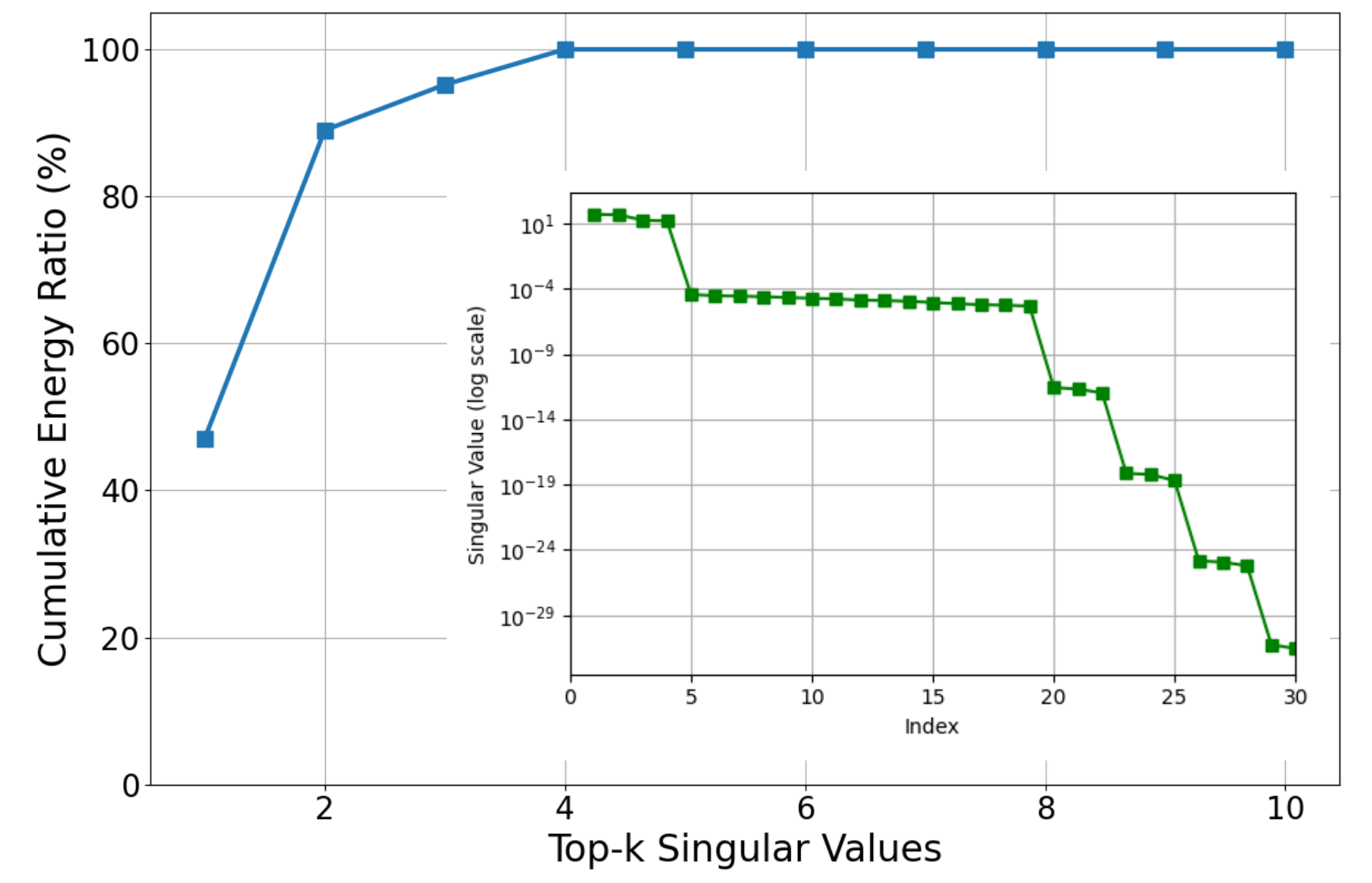} 
\caption{
The singular values of the prompt matrix with a length of 12 and the corresponding cumulative energy ratio.
}
\label{Fig:rankvalue}
\end{figure}

Additionally, to further validate the rationale for setting the rank to 4, we visualized in Figure \ref{Fig:rankvalue} the results of the singular value decomposition (SVD) along the feature dimension of the prompt vectors when no low-rank decomposition is applied during training and the prompt length is set to 12, as well as the corresponding cumulative energy ratio. It is can be observed that the cumulative energy of the top 4 singular values approaches 100\%, indicating that the effective rank of prompt matrix is 4. This observation is consistent with our experimental results.

\noindent\textbf{Explicit interaction enhanced prompts.} 
To validate the effectiveness of the proposed explicit interaction enhancement, we visualized the t-SNE distributions of the intermediate layer output feature maps in the network. The comparison between (b) and (c) in Figure \ref{fig:tsne} clearly demonstrates that prompts effectively help the model distinguish different tasks, validating that prompts can activate the model's ability to handle specific tasks. The comparison between (c) and (d) in Figure \ref{fig:tsne} reveals that the intermediate features with the enhanced prompts align more closely with the t-SNE distribution among the degradations shown in Figure \ref{fig:tsne}(a), which validates the effectiveness of the proposed contrastive constraint. Moreover, Table \ref{tab:ablation} further support our viewpoint, as the explicitly enhanced prompts have improved the performance over the model that only incorporates unenhanced prompts.

\begin{figure}[t]
\centering
\includegraphics[width=0.95\linewidth]{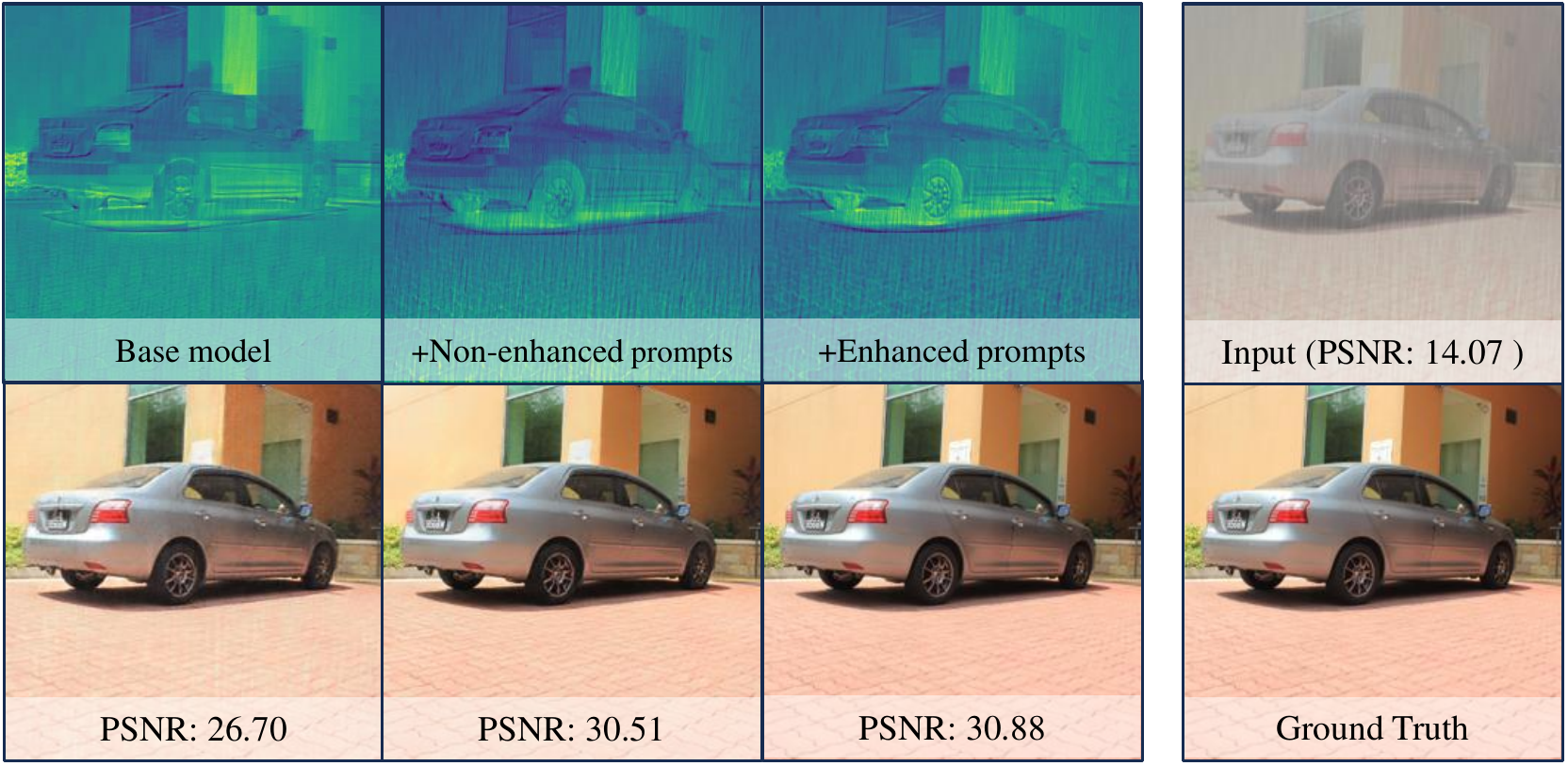} 
\caption{
Visualization of internal self-attention maps of models under different strategy enhancements. 
}
\label{Fig:attn-map}
\end{figure}

\noindent\textbf{How task-aware enhanced prompts adapt the model.} We visualize the changes in the model’s internal self-attention maps and corresponding outputs before and after adding prompts in Figure \ref{Fig:attn-map}. It is evident that the attention map after adding non-enhanced prompts evolves noticeably towards a degradation artifact distribution, with a more pronounced effect after using enhanced prompts. This exposes that interaction-enhanced prompts effectively help the model correctly identify the degradation artifacts, thereby improving the model's performance.

\begin{table}[t]
\centering
\caption{Ablation study on different settings. 'Base' refers to the pretrained model, '$+ P_{full}$' denotes the original Prompting Strategy, '$+ P_{attn}$' denotes the proposed Attention-level Prompting Strategy, and '$+ P_{attn-E}$' indicates the model with Task-aware Enhanced Prompts. '*' indicates joint training.}
\label{tab:ablation}
\scalebox{0.74}{
\begin{tabular}{l|c|c|c|c|c}
\toprule
\textbf{Method} & \textbf{Derain} & \textbf{Desnow} & \textbf{Dehaze} & \textbf{Raindrop} & \textbf{Avg.} \\
\midrule
Base & 28.71 & 28.53 & 31.26 & 29.27 & 29.44 \\
$+ P_{full}$ & 29.18 \textcolor{green}{(+0.47)} & 29.23 \textcolor{green}{(+0.70)} & 31.55 \textcolor{green}{(+0.29)} & 31.22 \textcolor{green}{(+1.95)} & 30.05 \textcolor{green}{(+0.61)} \\
$+ P_{attn}*$ & 29.37 \textcolor{green}{(+0.66)} & 29.55 \textcolor{green}{(+1.02)} & 31.29 \textcolor{green}{(+0.03)} & 32.64 \textcolor{green}{(+3.37)} & 30.71 \textcolor{green}{(+1.27)} \\
$+ P_{attn}$ & 30.03 \textcolor{green}{(+1.32)} & 29.89 \textcolor{green}{(+1.36)} & 31.88 \textcolor{green}{(+0.62)} & 31.62 \textcolor{green}{(+2.35)} & 30.86 \textcolor{green}{(+1.42)} \\
$+ P_{attn-E}$ & 30.18 \textcolor{green}{(+1.47)} & 30.47 \textcolor{green}{(+1.94)} & 32.77 \textcolor{green}{(+1.51)} & 32.13 \textcolor{green}{(+2.86)} & 31.39 \textcolor{green}{(+1.95)} \\
\bottomrule
\end{tabular}
}
\end{table}

\begin{table}[t]
    \centering
    \caption{Ablation Study on different prompt settings. We report the average performance of the model with prompts of different ranks and lengths across four tasks.}
    \label{tab:ablationforrank}
    \scalebox{0.75}{
        \begin{tabular}{c|ccccc}
        \toprule
        \textbf{Prompt Setting} & \multicolumn{5}{c}{\textbf{Rank}} \\
        \midrule
        {\bf Length} & \textbf{0} & \textbf{4} & \textbf{8} &\textbf{12} & \textbf{16}\\
        \midrule
        \textbf{0}  & \cellcolor[HTML]{E7E6E6} 29.443 & - & - & - & - \\
        \textbf{4}  & 29.690 & \cellcolor[HTML]{E7E6E6} 29.838& - & - & - \\
        \textbf{8}  & 29.885 & 30.243 & \cellcolor[HTML]{E7E6E6} 30.143& - & -  \\
        \textbf{12} & 29.458 & \underline{\bf 30.855} & 30.517 & \cellcolor[HTML]{D0CECE} 30.381 & - \\
        \textbf{16} & 29.835 & 30.153 & 30.520 & 30.100 & \cellcolor[HTML]{E7E6E6} 30.394\\
        \bottomrule
        \end{tabular}
    }
\end{table}

\subsection{Training Strategy and Hyperparameters}
\label{sec:Ablation-train}
\noindent \textbf{Training Strategy.}
According to Table \ref{tab:ablation}, joint training of the prompts and the backbone ($+ P_{attn}*$) results in an average performance gain that is 0.015 (0.127 vs. 0.142) lower compared to separate training, suggesting that task conflicts introduced by joint training lead to suboptimal performance.

\noindent \textbf{Hyperparameters.}
We follow \cite{transweather} settings for $\lambda_{per}$ and $S$ during pretraining. We also conduct experiments applying L1 loss and found that the former resulted in an average PSNR improvement of approximately 0.7 across all tasks. For the value of $\lambda_{cont}$, we perform a grid search and find 0.1 to be the optimal choice. Details are shown in supplementary materials.

\section{Conclusion}
In this work, we propose a novel AiOIR framework, \textit{\textbf{TAP}}. We employ task-aware enhanced prompts to strike a balance between task modeling and parameter efficiency. 
Specifically, we enhance prompt adaptability through both implicit and explicit interactions.
In addition, a two-stage training strategy is adopted to mitigate task conflicts. With only 2.75M parameters, our method achieves SOTA performance across various adverse weather removal tasks.

\section*{Acknowledgements}
This work was supported by the 'Pioneer' and 'Leading Goose' R\&D Program of Zhejiang under (Grant No. 2025C02110), Public Welfare Research Program of Ningbo under (Grant No. 2024S062), and Yongjiang Talent Project of Ningbo under (Grant No. 2024A-161-G).

\printbibliography

\end{document}